\documentclass[10pt,twocolumn,letterpaper]{article}

\usepackage{cvpr}
\usepackage{times}
\usepackage{epsfig}
\usepackage{graphicx}
\usepackage{amsmath}
\usepackage{amssymb}
\usepackage{lipsum}
\usepackage{float}

\usepackage[breaklinks=true,bookmarks=false]{hyperref}

\usepackage[capitalize]{cleveref}

\cvprfinalcopy 

\def\cvprPaperID{****} 

\begin{document}

\title{Color Constancy by GANs: An Experimental Survey}

\author{Partha Das$^{1,2}$
\and Anil S. Baslamisli$^1$
\and Yang Liu$^{1,2}$
\and Sezer Karaoglu$^{1,2}$
\and Theo Gevers$^{1,2}$
\and $^1$Computer Vision Lab, University of Amsterdam 
\and $^{2}$3DUniversum
\and {\tt\small \{p.das, a.s.baslamisli, th.gevers\}@uva.nl, \{yang, s.karaoglu\}@3duniversum.com}
}

\maketitle

\begin{abstract}
In this paper, we formulate the color constancy task as an image-to-image translation problem using GANs. By conducting a large set of experiments on different datasets, an experimental survey is provided on the use of different types of GANs to solve for color constancy i.e. CC-GANs (Color Constancy GANs). Based on the experimental review, recommendations are given for the design of CC-GAN  architectures based on different criteria, circumstances and datasets.
\end{abstract}

\section{Introduction}

\begin{figure}[t]
 \centering
  \label{figure1}
  \begin{tabular}{cc}
    \includegraphics[width=0.20\textwidth]{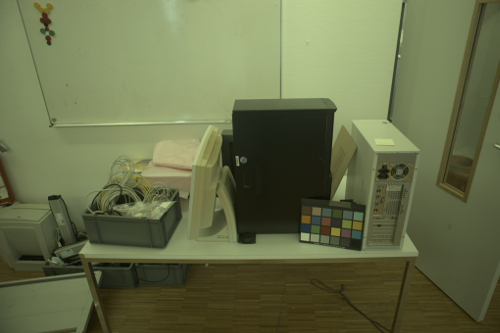}&
    \includegraphics[width=0.20\textwidth]{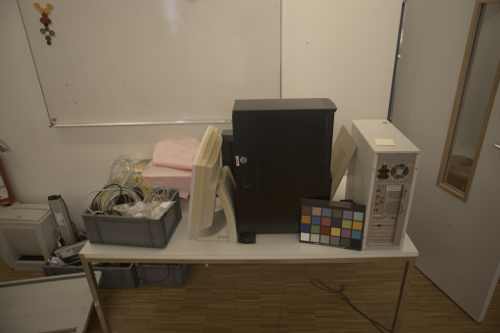}\\
(a) & (b) \\
   \includegraphics[width=0.20\textwidth]{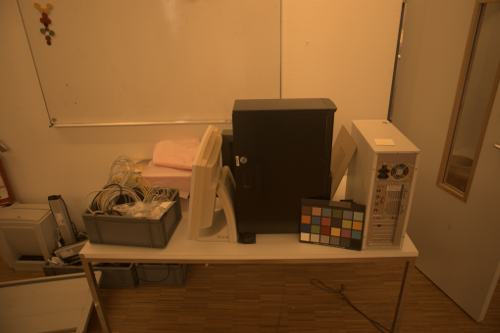}&
    \includegraphics[width=0.20\textwidth]{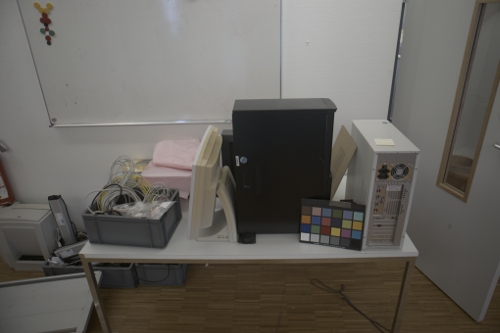}\\
 (c) & (d)\\
  \end{tabular}
  \vspace{+1mm}
\caption{(a) is an image taken under an unknown light source, (b) and (c) are images of the same scene under \emph{Illuminant D50} and \emph{Illuminant A} respectively. The task of color constancy is to recover the white-balanced image (d) from images (a), (b) or (c).}
\end{figure}

The observed colors in a scene are determined by object properties (albedo and geometry) and the color of the light source. The human visual system has, to a large extent, the ability to perceive object colors invariant to the color of the light source. This ability is called color constancy (CC), or auto white balance (AWB), as illustrated in Figure~\ref{figure1}. Although the ability of color constancy is trivial for humans, it's a difficult problem in computer vision because, for a given image, both the spectral reflectance (albedo) and the power distribution (light) are unknown.

Therefore, traditional color constancy algorithms impose priors to estimate the color of the light source. For example, the grey world algorithm~\cite{Buchsbaum1980} assumes that the average reflectance of the surfaces in a scene is achromatic under a (white) canonical light source. Then, using a diagonal model (i.e. von Kries model~\cite{Kries1970}), the image can be corrected such that it appears as if it was taken under a white light source. On the other hand, motivated by the success of convolutional neural networks (CNNs) obtained for other computer vision tasks, more recent work uses these powerful deep CNN models. For example,~\cite{Bianco2015,Lou2015} are the first to use CNNs for illuminant estimation and achieving state-of-the-art results.

In parallel, image-to-image translation tasks gained attention by the introduction of conditional generative adversarial networks (GANs)~\cite{Mirza2016}. The goal is to map one representation of a scene into another. For example, converting day images to night or transferring images into different styles. Our aim is to model the color constancy task as an image-to-image translation problem where the input is an \emph{RGB} image of a scene taken under an unknown light source and the output is the color corrected (white balanced) one.

Therefore, in this paper, we propose to formulate the color constancy task as an image-to-image translation problem using GANs. By conducting a large set of experiments on different datasets, we provide an experimental survey on the use of different types of GANs to solve the color constancy problem. The survey includes an analysis of the performance, behavior, and the effectiveness of the different GANs. The image-to-image translation problem for color constancy is defined as follows: the input domain is a set of regular \emph{RGB} images of a scene taken under an unknown light source and the output domain is the set of corresponding color corrected images for a canonical (white) light source. The following state-of-art image-to-image translation GANs are reviewed: \emph{Pix2Pix}~\cite{Isola2017}, \emph{CycleGAN}~\cite{Zhu2017}, and \emph{StarGAN}~\cite{Choi2018}. The goal is to analyze the performance of the different GAN architectures applied on the illumination transfer task. We evaluate the different GAN architectures based on their design properties, scene types and their generalization capabilities across different datasets. Finally, as a result of the experimental review, recommendations are given for the design of new GAN architectures based on different criteria, circumstances and datasets.

In summary, our contributions are: (1) to the best of our knowledge, we are the first to formulate the color constancy task as an image-to-image translation problem using GANs, (2) we provide extensive experiments on 3 different state-of-the-art GAN architectures to demonstrate their (i) effectiveness on the task and (ii) generalization ability across different datasets, and (3) we provide a thorough analysis of possible error contributing factors to aid future color constancy research.

\section{Related Work}

 \textbf{Color Constancy:} To mimic the human ability of color constancy for cameras, a wide range of algorithms are proposed. These algorithms can be classified into two main categories: (1) static-based methods which are either relying on reflection models or low-level image statistics, and (2) learning-based methods which rely on training sets~\cite{Gijsenij2011}. 
 
 Static methods assume that an image that is taken under a canonical light source  follow certain properties. For example, the grey-world algorithm~\cite{Buchsbaum1980} assumes that the average reflectance of the surfaces in a scene is achromatic. Further, the white-patch algorithm~\cite{Land1977} assumes that the maximum pixel response in an image corresponds to white reflectance. These assumptions are extended to first- and higher-order image statistics to provide more accurate estimators such as the grey-edge assumption~\cite{Weijer2007}. In their work,~\cite{Weijer2007} proves that all the statistics-based methods can be unified into a single framework. 

Learning-based methods require a set of labeled images to train a color constancy model. Since different statistics-based methods are suited for different scenes, algorithms are proposed to learn how to select (or fuse) the color constancy algorithm which works best for a given scene. For example,~\cite{Bianco2008} first classifies an image into an indoor or outdoor scene. Then, it determines the most appropriate algorithm for that scene.~\cite{Gijsenij2010} uses Weibull parametrization, ~\cite{Weijer2007-2} relies on high-level semantic features, and~\cite{Bianco2010} uses decision forests with various low-level features. In addition,~\cite{Rosenberg2004} proposes to estimate the posterior of the illuminant conditioned on surface reflectance in a Bayesian framework. Instead of modelling the distribution explicitly, the exemplar-based method~\cite{Joze2014} directly searches for the nearest patches in the training set to adapt a voting scheme for the final illuminant estimation. Furthermore, gamut-based methods~\cite{Barnard2000,Forsyth1990,Gijsenij2010-2} assume that only a limited number of colors can be observed in a scene. First, the canonical gamut is computed from the training set as a reference. Then, the input gamut is mapped onto the canonical gamut such that it completely fits in. Finally, the mapping is used to compute the corrected image. Moreover,~\cite{Agarwal2006,Xiong2006} use regression models, whereas~\cite{Cardei2002,Stanikunas2004} use neural networks.

More recent work uses powerful deep CNN models to approach the color constancy problem.~\cite{Lou2015} uses an end-to-end deep model to estimate the illuminant.~\cite{Bianco2015} extracts patch features from a pre-trained CNN and feed them to a regression module to estimate the scene illumination. Both works provide state-of-the-art results supported by powerful CNN models. In another work,~\cite{Oh2017} formulates the problem as an illumination classification problem using deep learning.~\cite{Shi2016} proposes a two-branch deep CNN. One generates multiple hypotheses of the illuminant of a given patch. The other branch adaptively selects the best hypothesis. Moreover,~\cite{Bianco2017} estimates the local illuminant color of image patches.~\cite{Hu2017} proposes a method to weight the input patches to differentiate between semantically valuable and semantically ambiguous local regions. Finally,~\cite{Qian2017} exploits multiple frames over time within a deep learning framework. On the other hand,~\cite{Yan2018} bypasses the illuminant color estimation step and directly learns a mapping from the input image to the white balanced image in an end-to-end fashion. 

 \textbf{Generative Adversarial Networks:} Recently, generative adversarial networks (GANs)~\cite{Goodfellow2014} gained a lot of attention as they are successful in various computer vision tasks. The network adversarially learns to generate images that are very close to real image samples. They have demonstrated remarkable progress in the fields of image translation~\cite{Isola2017,Kim2017,Zhu2017}, image super-resolution~\cite{Ledig2017}, and image synthesis~\cite{Choi2018,Isola2017,Zhang2017}.
 
 Traditional GANs use random noise as input to generate images. However, noise inputs do not have much control over the generation. Further improvements~\cite{Choi2018,Kim2017,Mirza2016,Reed2016} show how to constrain the generated outputs by conditioning on the input with specific attributes. By conditioning on the input image,~\cite{Kim2017} is able to generate objects across domains, like handbags to shoes. \emph{Pix2Pix}~\cite{Isola2017} introduces a pixel-wise paired generation for an image-to-image translation task from paired images. The authors illustrate its success on several translation tasks such as transferring from edges to objects or facades to buildings.~\cite{Liu2017,Zhu2017} present an improvement upon this work by learning the attributes of the input and output domains, and training the network in an unsupervised manner. Specifically,~\cite{Zhu2017} introduces a cycle consistency through an architecture called \emph{CycleGAN}, which ensures that the generated images are mapped back to the original image space. In this context, we propose to model color constancy as an image-to-image translation task where the source domain is a camera sensor image and the target domain is the white balanced version. This enables to map back the white balanced version to the sensor domain by cycle consistency. Although these methods approach the problem with unpaired data, they are only able to achieve one domain transfer at a time. Therefore,~\cite{Choi2018} proposes \emph{StarGAN}, to learn multiple attributes across domains simultaneously. This is then applied to faces where the model can be steered to generate different outputs based on conditioning vectors. For example, transferring from blond to black hair or making someone look younger or older. This can also be applied to the color constancy task as different illuminants can be learned by providing a conditioning vector. In this way, various color corrected images can be generated. This can be beneficial to render the same scene under different illuminant colors or do a consistency check to achieve the same canonical light under different illuminant colors. In the following section, we give an overview of the problem of color constancy, followed by the GAN architectures used in our survey. 
 
 \begin{figure*}[h]
    \centering
    \includegraphics[scale=0.5]{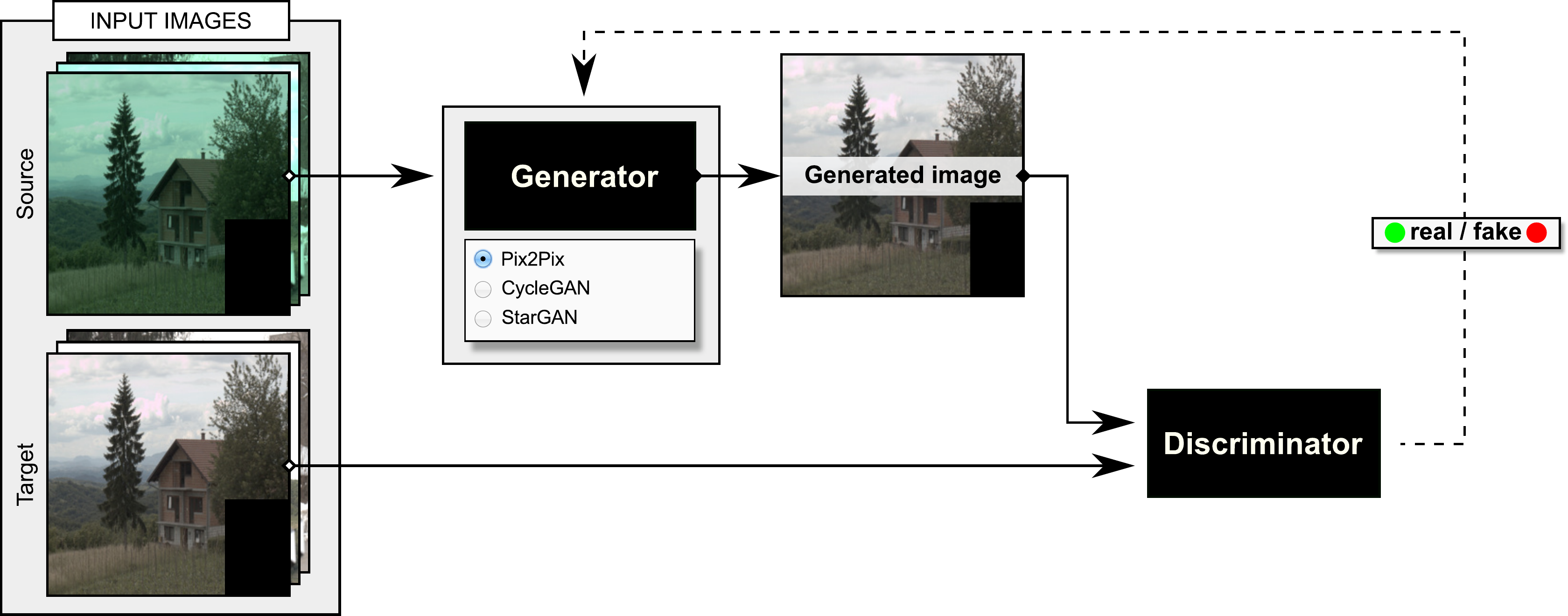}
    \caption{Overview of GAN-based color constancy estimation. The input image is provided as a condition to the Generator ($G$), whose output, along with a true sample is passed to the Discriminator ($D_Y$).}
    \label{fig:network}
\end{figure*}
 
\section{Methodology}
\subsection{Color Constancy}
For a Lambertian (diffuse) scene, the \emph{RGB} values of an image \emph{I} taken by a digital camera are determined by object properties (albedo $\rho(\lambda)$ and geometry $m$), spectral power distribution (color) of the illuminant $e(\lambda)$ and camera spectral sensitivity function  $f(\lambda)$ as follows~\cite{Shafer1985}:

\begin{equation} \label{eq:imf}
\begin{aligned}
I = m(\vec{n}, \vec{s}) \int_{\omega}^{} \rho(\lambda)\; e(\lambda)\;f(\lambda)\; \mathrm{d}\lambda \;,
\end{aligned}
\end{equation}

\noindent where $\lambda$ denotes the wavelength of the incoming light, $\omega$ is the visible spectrum, $\vec{n}$ indicates the surface normal, and $\vec{s}$ represents the light source direction. 

Color constancy aims to correct an image such that it appears as if it was taken under a canonical (white) light source. First, the color of the light source is estimated. Then, the image is transformed into an image observed under a reference (white) light source. The correction process can be regarded as chromatic adaptation~\cite{Fairchild2013} in which the problem is formulated as a linear transformation: 

\begin{equation} \label{eq:canon}
    I = W * e\;,    
\end{equation}

\noindent where \emph{W} denotes the canonical image in linear $RGB$ space (white balanced image) and \emph{e} is the color of the illuminant. The von Kries model~\cite{Kries1970} is used as the transformation function to recover the color of the light source per pixel:
\begin{equation} \label{eq:vonKries}
\begin{aligned}
\begin{pmatrix}
R_{c} \\
G_{c} \\
B_{c}
\end{pmatrix} =
\begin{pmatrix}
e_{R} & 0 & 0 \\
0 & e_{G} & 0 \\
0 & 0 & e_{B}
\end{pmatrix}
\begin{pmatrix}
R_{u} \\
G_{u} \\
B_{u}
\end{pmatrix}\;,
\end{aligned}
\end{equation}

\noindent where \emph{u} denotes the unknown light source and \emph{c} represents the canonical (corrected) illuminant. The aim is to recover $e = (e_{R}, e_{G}, e_{B})^{T}$. The model describes the color constancy as a problem of modifying the gains of the image channels independently. Under this model, given an image from a camera sensor captured under non-canonical lighting conditions, the problem of color constancy is defined as the estimation of \emph{e} given \emph{I}. Thus, it estimates the image under a canonical light source.

Given that we have 3 variables $(I, W, e)$ of which only 1 $(I)$ is known, the problem is under constrained. Traditional work usually estimates the illuminant color fist, and then recovers the white balanced image. In this work, we model the color constancy task as an image-to-image translation problem, and directly generate the white balanced image by generative adversarial networks (GANs). 

\subsection{Generative Adversarial Networks (GANs)}

A GAN consists of a generator $G$ and a discriminator pair $D_{Y}$. As shown in Figure~\ref{fig:network}, $G$ takes as input $X$, an $RGB$ image from the camera response under unknown light source, and outputs $Y$, the same image under a canonical light source. We want to learn a mapping $G: X \rightarrow Y$ from a domain having an unknown light source to a domain having a reference (white) light source without modifying the pixel intensities or scene structure. The output is then passed to $D_{Y}$ which classifies whether the input is sampled from the real color corrected domain or is generated by $G$. The supervision signal provided by $D_{Y}$ forces $G$ to learn to transform the input with unknown light to appear closer to the color corrected images. In this paper, we select $3$ representative state-of-the-art GAN architectures for the color constancy problem.

\textbf{Pix2Pix~\cite{Isola2017}:} The model uses a U-Net~\cite{Ronneberger2015} architecture as the generator. This allows the architecture to learn an image transformation, conditioned on the input image. Skip connections~\cite{Mao2016} are used to transfer high frequency details from encoder to the decoder blocks. The discriminator is fully convolutional $PatchGAN$, which evaluates the generated samples at the scale of (overlapping) image patches. It enforces the high frequency consistency by modelling the image as a Markov random field.

The network is trained with an adversarial loss:
\begin{equation}\label{eq:GANLoss}
    \begin{split}
        \mathcal{L}_{GAN}(G, D_Y, X, Y) &= E_{y \sim p_{data}(y)}[log D_Y(y)]\;+\\
                            &E_{x \sim p_{data}(x)}[log(1 - D_Y(G(x)))]\;,
    \end{split}
\end{equation}
\noindent where, $X$ is an $RGB$ image from the sensor domain under unknown light, and $Y$ is an image under canonical light. The discriminator $D_Y$ takes as input the image generated by the generator $G$ and predicts whether it was generated $G(x)$ or coming from real samples $y$. Additionally, an extra L1 loss is used to produce generated images to be close to the ground-truth output:
\begin{equation}\label{eq:l1loss}
    \mathcal{L}_{L1}(G) = E_{x, y}||y - G(x)||_1^{1}\;,
\end{equation}
Thus, the final combined loss becomes:
\begin{equation}
    L_{pix2pix} = \mathcal{L}_{GAN}(G, D_Y, X, Y) + \mathcal{L}_{L1}(G)\;.
\end{equation}
The pixel-wise correspondence is particularly suited to the color constancy problem as the white-balanced image is a pixel-wise scaling of the color image. The overall structure remains the same. An advantageous aspect of the method for the problem of color constancy is that this approach, unlike  traditional methods, learns a local transformation rather than a global transformation. This is useful when the scene illumination is non-uniform. A disadvantage is that the model requires paired images for training.

\textbf{CycleGAN~\cite{Zhu2017}:} To discard the dependency on paired image data, we adopt \emph{CycleGAN} operating on unpaired data. It uses samples from two independent distributions. Modelling the learning process from unpaired data allows the network to learn a global transformation. This makes the model more resilient to biases. The generator uses a ResNet~\cite{He2016} based architecture. Following~\cite{Isola2017}, the discriminator is a \emph{PatchGAN}. To train the network in an unpaired way, CycleGAN introduces a cycle consistency loss:
\begin{equation}\label{eq:cycle}
    \mathcal{L}_{cycle}(G,F) = E_{x \sim p_{data}(x)}\;||F(G(x)) - x||_{1}^{1}\;,
\end{equation}
\noindent where $G$ and $F$ are functions parameterized by an encoder-decoder network that learns a transformation of an image from one domain to another. It is similar to Equation~\ref{eq:l1loss}. However, in this case, instead of using the output directly from the generator $G$, the output is passed through another generator $F$ that learns to translate the output back to the original input domain. This is termed as a forward cycle, where the transformation is defined as $X \rightarrow Y$. Similarly, a second backward cycle is introduced for the transformation $Y \rightarrow X$, to enforce a full cycle and prevent the network from learning the delta function over the distribution. 

In our case, $G$ learns a translation from the sensor domain with unknown light to the white balanced domain, and $F$ learns the vice versa. Then, the model using cycle consistency combines the adversarial and cycle loss as follows:
\begin{equation}\label{eq:cycleLossFinal}
    L_{cycleGAN} = \mathcal{L}_{GAN}(G, D_Y, X, Y) + 10 * \mathcal{L}_{cycle}(G,F) \;.
\end{equation}
This allows the network to be independent of the local pixel-wise dependence. The network learns a global transformation for the input image to the white balanced domain. Due to the unpaired image domains, one could simply use arbitrary data to build the sensor space distribution and correspondingly curate professionally white balanced images to train a network. However, a problem arises when the requirement is to have multiple light sources. In such case, one would need to train a separate model for each of the illuminant transformation combinations.

\textbf{StarGAN~\cite{Choi2018}:} This architecture specifically addresses the issue of one-to-one transformation learning. Using this architecture, it is possible to learn one-to-many mappings, which is the limitation of \emph{CycleGAN}. The model builds upon the work of \emph{CycleGAN}, in which the generator uses an additional condition in the form of a domain vector. The vector specifies which target domain the network output should resemble. Similar to CycleGAN, a second generator is used to enforce cycle consistency, Equation~\ref{eq:cycle}. The discriminator is also modified to accommodate one-to-many mappings. It not only discriminates on the image whether being from the real or generated distribution, but also learns to classify the input image domain. This way, the discriminator is able to provide meaningful supervision signals to the generator, not only steering it to the real distribution, but also making sure it follows a certain domain distribution. As a result, the final loss is a combination of Equation~\ref{eq:cycleLossFinal} and the domain classification loss.

The ability to define domains allows us to train a network that is able to learn a global mapping for different illuminants (i.e. multiple light sources) by a single model. Similar to CycleGAN, the model does not require paired data. It is possible to train a model to learn a mapping between the various domains when sufficient amount of data is available. Furthermore, since the model is able to learn multi-domain transformations, it is able to learn a common transformation that is capable of mapping the camera sensor input image to multiple illuminant types. In this way, we can render the same scene under different illuminant colors or do a consistency check to see if the model can achieve the same canonical light under different illuminant colors. 

For the experiments, StarGAN is trained to estimate both the \emph{canonical} transformation and \emph{Illuminant A}. That translates to 3 different domains; the canonical, Illuminant A and the input domain. Since the model learns in an unsupervised manner, StarGAN also learns the representation for the input domain as a valid target domain.

\subsection{Illuminant Color Prediction}
For the experiments, instead of first predicting the color of the light source and then correcting the image, we directly estimate the white balanced image. Hence, the single illuminant color $e$ is estimated by calculating the inverse of Equation~\ref{eq:canon}:

\begin{equation} \label{eq:metric}
    e = I * \hat{W} ^{-1}\;,   
\end{equation}

\noindent where $I$ is the input image and $\hat{W}$ is the white balanced estimation from the network. Both $I$ and $\hat{W}$ are converted into the linear $RGB$ space before computing the illuminant.

\section{Experiments}

\subsection{Datasets}

We evaluate the performance of the models on 3 standard color constancy benchmarks. The SFU Gray Ball~\cite{Ciurea2003} dataset contains over 11K images captured by a video camera. A gray ball is mounted on the camera to calculate the ground-truth. We use the linear ground-truth for evaluations. Second, we use ColorChecker dataset~\cite{Shi0000} that contains 568 linear raw images including indoor and outdoor scenes. A Macbeth color checker is placed in every scene to obtain the ground truth. Recently,~\cite{Finlayson2017} pointed out that original dataset has 3 different ground-truths and for each ground-truth the performance of various color constancy algorithms are different. Instead of using~\cite{Shi0000}, we use the recalculated version of~\cite{Finlayson2017} called \emph{ColorChecker RECommended} (RCC). Finally,~\cite{Banic2018} provides a dataset of 1365 outdoor images with a cube placed in every scene. Two different ground-truths are estimated from the two gray faces of the cube in different directions. Finally, one is considered as the final ground-truth by visual inspection. For the experiments, the datasets are randomly split into 80\% training and 20\% testing. The reference objects present in the images are masked out to prevent networks from cheating (directly estimating the color of the light source from the reference).

\subsection{Error Metric}
Following the common practice, we report on the angular error between the ground-truth illuminant $RGB$ $e$ and the estimated illuminant $RGB$ $\hat{e}$: 

\begin{equation} \label{eq:angularError}
    d_{ang} = arccos(\dfrac{e \cdot \hat{e}}{||e|| \times ||\hat{e}||})\;,
\end{equation}

\noindent where $||.||$ is the L2 norm. For each image, the angular error is computed, and the mean, median, trimean, means of the lowest-error 25\% and the highest-error 25\%, and maximum angular errors are reported to evaluate the models quantitatively.

\section{Evaluation}

\subsection{Linear RGB vs. sRGB}
In this experiment, we evaluate the performance of the architectures based on the input type; linear $RGB$ vs. $sRGB$. Traditional work usually models the color constancy as a problem of modifying the gains of the image channels independently. As a result, they generally operate on the linear $RGB$ color space. However, in the case of GANs, that assumption is no longer a requirement as they can implicitly estimate a transformation matrix that encodes such information. Moreover, for the linear space, the range of pixel values is often compressed to one part. In practice, 12 bit sensor images are encoded in 16 bits. This approach handles the problem of clipped values and saturated pixels, which will otherwise may bias the results towards x1.0 white balance gains. Nonetheless, the process also creates 4 trailing zero bits. On the other hand, in $sRGB$ space, the distribution of the pixel values are more uniform. The color distribution of the linear image is close to zero. Therefore, we report a performance comparison between the linear $RGB$ color space and the $sRGB$ color space by using \emph{CycleGAN}~\cite{Zhu2017}. The model directly outputs a white balanced estimation of an input image. Then, the color of the light source is estimated by Equation~\eqref{eq:metric}. The results are presented in~\cref{tab:lin_ball,tab:lin_cc2,tab:lin_cube} for the different datasets. 

\begin{table}[h]
\centering
\scalebox{0.8}{
\begin{tabular}{|l|c|c|c|c|c|c|}
\hline
Input & Mean & Med. & Tri. & Best 25\% & Worst 25\% & Max \\
\hline \hline
Linear $RGB$ & 9.2 & \textbf{5.7} & 6.9 & 1.6 & 21.6 & 39.3\\ \hline
$sRGB$ & \textbf{8.4} & 5.9 & \textbf{6.4} & \textbf{1.5} & \textbf{19.6} & \textbf{37.8}\\
\hline
\end{tabular}}
\vspace{+1mm}
\caption{Performance comparison for linear $RGB$ versus $sRGB$ color spaces for the SFU Gray Ball dataset~\cite{Ciurea2003}. $sRGB$ input yields better results.}
\label{tab:lin_ball}
\end{table}

\begin{table}[h]
\centering
\scalebox{0.8}{
\begin{tabular}{|l|c|c|c|c|c|c|}
\hline
Input & Mean & Med. & Tri. & Best 25\% & Worst 25\% & Max \\
\hline \hline
Linear $RGB$ & 5.8 & 5.5 & 5.4 & 2.4 & 10.0 & \textbf{15.9}\\ \hline
$sRGB$ & \textbf{3.4} & \textbf{2.6} & \textbf{2.8} & \textbf{0.7} & \textbf{7.3} & 18.0\\
\hline
\end{tabular}}
\vspace{+1mm}
\caption{Performance comparison for linear $RGB$ versus $sRGB$ color spaces for the ColorChecker RECommended dataset~\cite{Finlayson2017}. $sRGB$ input yields better results.}
\label{tab:lin_cc2}
\end{table}

\begin{table}[h]
\centering
\scalebox{0.8}{
\begin{tabular}{|l|c|c|c|c|c|c|}
\hline
Input & Mean & Med. & Tri. & Best 25\% & Worst 25\% & Max \\
\hline \hline
Linear $RGB$ & 6.2 & 5.8 & 5.8 & 2.7 & 10.4 & 20.5\\ \hline
$sRGB$ & \textbf{1.5} & \textbf{1.2} & \textbf{1.3} & \textbf{0.5} & \textbf{3.0} & \textbf{6.0}\\
\hline
\end{tabular}}
\vspace{+1mm}
\caption{Performance comparison for linear $RGB$ versus $sRGB$ color spaces for Cube dataset~\cite{Banic2018}. $sRGB$ input yields better results.}
\label{tab:lin_cube}
\end{table}

The experiments show that the performance of $sRGB$ color space is better than linear $RGB$ for all datasets. As a results, all subsequent experiments are performed in the $sRGB$ color space to achieve maximum performance.  

\subsection{GAN Architecture Comparisons}

\begin{figure}[H]
    \centering
    \includegraphics[scale=0.35]{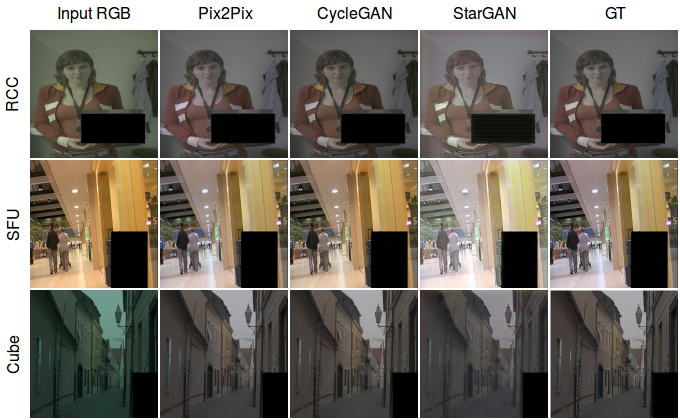}
    \caption{Performance of different GAN architectures on the color constancy task. All models are able to recover proper white balanced images. The images are gamma adjusted for visualization.}
    \label{fig:architectures}
\end{figure}

In this experiment, we compare the performance of 3 different state-of-the-art GAN architectures for color constancy; Pix2Pix~\cite{Isola2017}, CycleGAN~\cite{Zhu2017} and StarGAN~\cite{Choi2018}. All models use $sRGB$ inputs. Outputs are the white balanced estimates. Then, the color of the light source is estimated by Equation~\eqref{eq:metric}. The results are presented in~\cref{tab:gan_rec,tab:gan_cube,tab:gan_ball} for the different datasets.  Figure~\ref{fig:architectures} provides a number of visual examples.

\begin{table}[h]
\centering
\scalebox{0.75}{
\begin{tabular}{|l|c|c|c|c|c|c|}
\hline
Model & Mean & Med. & Tri. & Best 25\% & Worst 25\% & Max \\
\hline \hline
Pix2Pix~\cite{Isola2017} & \textbf{6.6} & \textbf{5.3} & \textbf{5.4} & \textbf{1.4} & \textbf{14.2} & \textbf{36.0}\\ \hline
StarGAN~\cite{Choi2018}& 10.3&8.9&9.0&4.0&18.9&46.0\\
\hline
CycleGAN~\cite{Zhu2017} & 8.4 & 5.9 & 6.4 & 1.5 & 19.6 & 37.8\\ 
\hline
\end{tabular}}
\vspace{+1mm}
\caption{Performance of different GAN models for SFU Gray Ball dataset~\cite{Ciurea2003}. Pix2Pix achieves the best performance.}\label{tab:gan_ball}
\end{table}

\begin{table}[h]
\centering
\scalebox{0.75}{
\begin{tabular}{|l|c|c|c|c|c|c|}
\hline
Model & Mean & Med. & Tri. & Best 25\% & Worst 25\% & Max \\
\hline \hline
Pix2Pix~\cite{Isola2017} & 3.6 & 2.8 & 3.1 & 1.2 & \textbf{7.2} & \textbf{11.3}\\ \hline
StarGAN~\cite{Choi2018} & 5.3&4.2&4.6&1.5&11.0&21.8\\ \hline
CycleGAN~\cite{Zhu2017} & \textbf{3.4} & \textbf{2.6} & \textbf{2.8} & \textbf{0.7} & 7.3 & 18.0\\
\hline
\end{tabular}}
\vspace{+1mm}
\caption{Performance of different GAN models for ColorChecker RECommended dataset~\cite{Finlayson2017}. CycleGAN achieves the best performance.}
\label{tab:gan_rec}
\end{table}

\begin{table}[h]
\centering
\scalebox{0.75}{
\begin{tabular}{|l|c|c|c|c|c|c|}
\hline
Model & Mean & Med. & Tri. & Best 25\% & Worst 25\% & Max \\
\hline \hline
Pix2Pix~\cite{Isola2017} & 1.9 & 1.4 & 1.5 & 0.7 & 4.0 & 8.0\\ \hline
StarGAN~\cite{Choi2018} & 3.8&3.3&3.5&1.3&7.0&11.4\\ \hline
CycleGAN~\cite{Zhu2017} & \textbf{1.5} & \textbf{1.2} & \textbf{1.3} & \textbf{0.5} & \textbf{3.0} & \textbf{6.0}\\
\hline
\end{tabular}}
\vspace{+1mm}
\caption{Performance of different GAN models for Cube dataset~\cite{Banic2018}. CycleGAN achieves the best performance.}
\label{tab:gan_cube}
\end{table}

\begin{figure}
    \centering
    \includegraphics[scale=0.45]{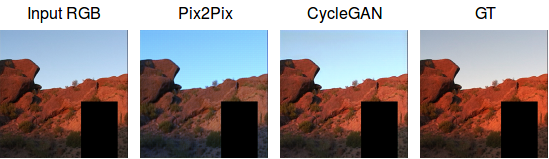}
    \caption{Resilience of Pix2Pix in a multi-illuminant case. The foreground is in shadow, while the background receives full illumination. This creates a multi-illuminant scene. Pix2Pix is able to recover the white-balanced image while keeping the foreground and background illuminant separate. CycleGAN estimates a global uniform lighting yielding a red cast. This is due to the inaccurate GT as the reference gray ball is placed in the shadowed foreground ignoring the multi-illuminant case.}
    \label{fig:failuregt}
\end{figure}

\cref{tab:gan_rec,tab:gan_cube} show that \emph{CycleGAN} outperforms \emph{Pix2Pix} and \emph{StarGAN}. Table~\ref{tab:gan_ball} shows that Pix2Pix achieves the best performance for the SFU Gray Ball dataset. Most of these cases are scenes with multiple light sources with strong lighting and shadowed regions. Pix2Pix learns a per-pixel mapping of the illumination, and is agnostic about the global illumination. This allows the model to effectively learn different local illumination conditions. Other GANs  learn a global representation of the scene illumination. This assumption only holds for uniform, global illumination. In the Gray Ball dataset, there are many instances of scenes with multiple light sources. Thus, Pix2Pix shows to be more resilience to multi-illuminant instances through its local mapping capability. Figure~\ref{fig:failuregt} shows an example of non-uniform illumination effects present in an image. Pix2Pix is more resilient as it can recover a proper white balanced image, while preserving the foreground and background (brightness) distinction. On the other hand, CycleGAN appears closer to the ground-truth. This can be explained by the fact that the gray ball that is used to estimate the ground-truth illuminant, is placed in the shadowed foreground. Hence, the estimation completely misses the multi-illuminant case. CycleGAN, due to its global estimator nature, estimates a uniform white balanced image which is closer to the ground-truth, but yields a red cast.

\begin{figure}
    \centering
    \includegraphics[scale=0.42]{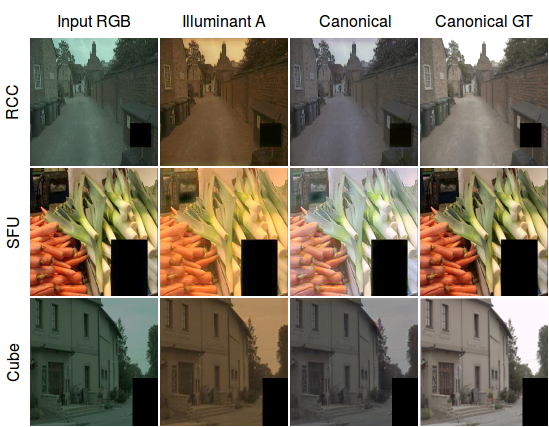}
    \caption{Visualizations of the StarGAN architecture. This architecture learns mappings to multiple domains, e.g. Illuminant A and canonical, simultaneously. The images are gamma adjusted for visualization.}
    \label{fig:stargan}
\end{figure}

StarGAN has much lower performance than the other 2 models. This may be due to the architecture's latent representation that has to compensate for learning 3 different (input, canonical, Illuminant A) domain transformations. This is not the case with either Pix2Pix or CycleGAN, where the mapping is one-to-one. We can make use of StarGAN outputting in multiple target illuminant domains, as visualized in Figure~\ref{fig:stargan}. This ability can also directly be applied to relighting tasks. The accuracy of this internal estimation defines how close the output is to the target domain. Ideally, this estimated ground-truth illumination should be the same for all the output domains. Hence, by checking the performance of the output in estimating the ground-truth using Equation~\ref{eq:metric}, we can check the consistency of the network. This also allows for observing which illuminant performs better in estimating the ground-truth illuminants. Per-illuminant consistency performance is presented in Table~\ref{tab:stargan_re_data}. It can observed from the table that for the RECommended Color Checker (RCC) dataset \emph{Illuminant A} estimation yields better results than the direct canonical estimation. For the Cube dataset, the canonical performs better. This could be explained as the Cube dataset consists of only outdoor images. Hence, the input images seldomly have an illuminant that is closer to  Illuminant A. Conversely, for the RCC dataset, indoor scenes generally have incandescent lighting, which is closer to Illuminant A. This intuition is further evaluated in the next experiment section where the performance of outdoor scenes are compared against indoors.

\begin{table}[h]
\centering
\scalebox{0.71}{
\begin{tabular}{|l|c|c|c|c|c|c|}
\hline
Dataset + Illumination & Mean  & Med. & Tri. & Best 25\% & Worst 25\% & Max \\
\hline \hline
RCC + Canonical & 5.3 & 4.2 & 4.6 & 1.5 & 11.0 & 21.8\\
RCC + Illuminant A & 2.9 & 2.2 & 2.3 & 0.8 & 6.4 & 13.8\\ \hline
Cube + Canonical & 3.8 & 3.3 & 3.5 & 1.3 & 7.0 & 11.4\\
Cube + Illuminant A & 4.1 & 3.2 & 3.4 & 1.1 & 8.5 & 20.7\\ \hline
SFU + Canonical & 10.3 & 8.9 & 9.0 & 4.0 & 18.9 & 46.0\\
SFU + Illuminant A & 12.7 & 11.2 & 11.2 & 3.2 & 24.8 & 61.1\\
\hline
\end{tabular}}
\vspace{+1mm}
\caption{Consistency of StarGAN's illumination estimation.}
\label{tab:stargan_re_data}
\end{table}

\subsection{Error Sources}

\begin{figure}
    \centering
    \includegraphics[scale=0.55]{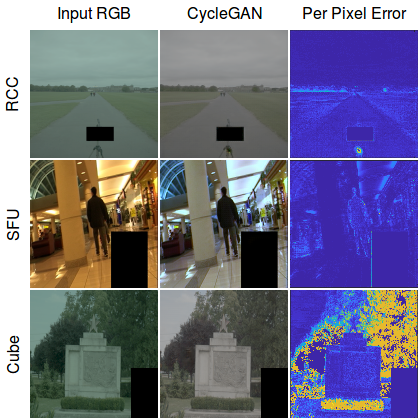}
    \caption{Pixel wise angular errors for the predicted images. Regions with smooth color changes and planar surfaces with homogeneous colors produce lower errors.}
    \label{fig:error_pics}
\end{figure}

Since our GAN-based approach bypasses the illumination estimation stage to directly predict the white balanced image, we can analyze the error sources of the generation process by calculating the per-pixel error image. The error images are generated by computing the per-pixel angular error between the estimated white balanced image and the input image corrected with the ground-truth illuminant. Figure~\ref{fig:error_pics} provides a number of examples of the estimations of the \emph{CycleGAN} model. It is shown that the regions with smooth color changes and planar surfaces with homogeneous colors produce lower errors. On the other hand, regions with complex textures and non-homogeneous colors contribute to higher errors. That is expected as the network learns to estimate white balanced images based on neighborhood relations. To overcome this, networks can be further regularized to give more attention to edges or surface normals. Further, the reference object of the RCC dataset causes high errors. Figure~\ref{fig:error_pics} shows that for the outdoor images of RCC and Cube, the input images have a greenish color cast. When the model predicts the white balanced images, green areas like grass or trees, whose color is closer to the color cast present in the input images, produce more errors. The reason may be that the networks confuses the color of the light source with object albedo. For the indoor scenes of SFU, most of the errors are caused by \emph{direct light} coming from the ceiling. 

\subsection{Cross Dataset Performance}
\begin{figure}[h]
    \centering
    \includegraphics[scale=0.55]{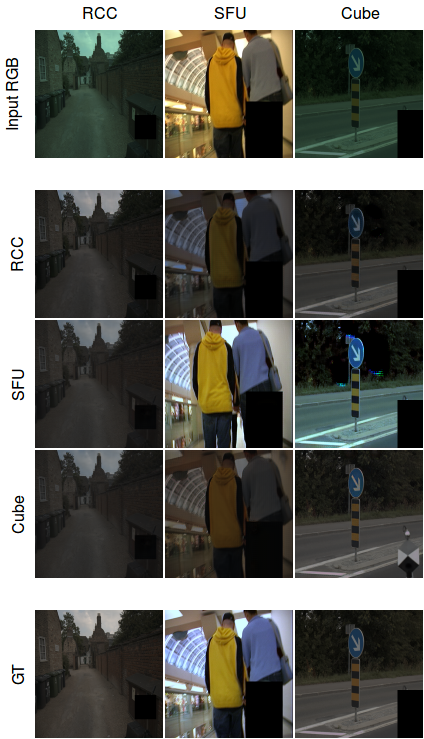}
    \caption{Cross dataset evaluations. Results on the SFU Gray Ball dataset appear much darker. Conversely, the result of the SFU Gray Ball trained model on the Cube and the RCC dataset have many saturation artifacts.}
    \label{fig:cross_datasets}
\end{figure}

Most of the learning-based illumination estimation tasks are usually trained and tested using the same datasets. This does not always give an intuition regarding the generalization capacity of the algorithm. In this experiment, we further evaluate the generalization ability of \emph{CycleGAN}, which achieves the best quantitative results over the majority of the 3 datasets. Since the datasets used for experiments have a variety of illumination conditions, picking the best performing model allows us to make an unbiased evaluation. Hence, we train CycleGAN on one dataset and test it on the other two. Same test splits are used for all experiments. The quantitative results are presented in Table~\ref{tab:cross_data} and visual results are shown in Figure~\ref{fig:cross_datasets}.  

\begin{table}[h]
\centering
\scalebox{0.71}{
\begin{tabular}{|l|c|c|c|c|c|c|}
\hline
Datasets (Train/Test) & Mean  & Med. &Tri. & Best 25\% & Worst 25\% & Max \\
\hline \hline
SFU / SFU & 8.4 & 5.9 & 6.4 & 1.5 & 19.6 & 37.8 \\
SFU / RCC & 10.9 & 10.5 & 10.7 & 5.6 & 16.6 & 23.5\\
SFU / Cube & 14.9 & 14.5 & 14.6 & 8.9 & 21.9 & 36.8\\ \hline
RCC / RCC & 3.4 & 2.6 & 2.8 & 0.7 & 7.3 & 18.0 \\
RCC / SFU & 10.1 & 9.3 & 9.0 & 2.5 & 20.2 & 45.5\\
RCC / Cube & 2.6 & 2.2 & 2.3 & 0.7 & 5.0 & 11.7\\ \hline
Cube / Cube & 1.5 & 1.2 & 1.3 & 0.5 & 3.0 & 6.0 \\
Cube / SFU & 11.6 & 10.4 & 10.5 & 3.1 & 22.2 & 41.4\\ 
Cube / RCC & 4.0 & 3.8 & 3.7 & 1.2 & 7.5 & 14.1\\
\hline
\end{tabular}}
\vspace{+1mm}
\caption{Generalization behavior of CycleGAN~\cite{Zhu2017}. Models trained on the RECommended Color Checker (RCC)~\cite{Finlayson2017} and Cube~\cite{Banic2018} datasets achieve the best generalization performance.}
\label{tab:cross_data}
\end{table}

Table~\ref{tab:cross_data} shows that the models trained on the RECommended Color Checker (RCC)~\cite{Finlayson2017} and Cube~\cite{Banic2018} datasets achieve the best generalization performance, while SFU Gray Ball dataset~\cite{Ciurea2003} has the highest errors. This can be attributed to the fact that the ground-truth for the dataset is not always accurate. The performance of the model trained on the RCC is better than the one trained on the Cube dataset. This is because the RCC dataset contains both indoor and outdoor scenes, and thus the model learns more variation. Figure~\ref{fig:cross_datasets} shows that the results on the SFU Gray Ball dataset appear darker. Conversely, the result of the SFU Gray Ball trained model on the Cube and the RCC dataset have many saturation artifacts. This is most likely due to the way the dataset is provided. Both RCC and Cube datasets are available as 16-bit images, which involves saturated pixel information. The SFU Gray Ball dataset is standard, camera processed (clipped) 8 bit images. Further, the model produces blurred results, therefore the gradient significantly differs from the original image.

\begin{figure}[h]
    \centering
    \includegraphics[scale=0.5]{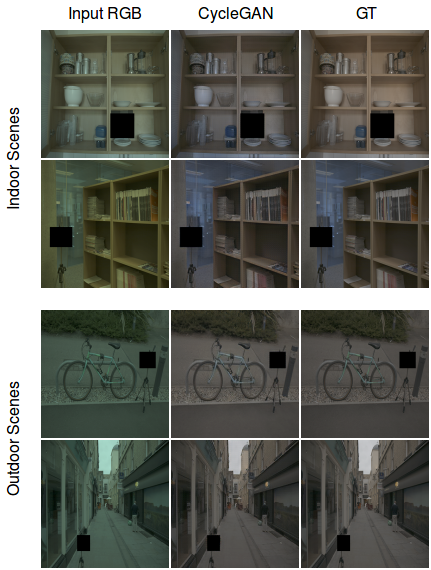}
    \caption{Visuals of scene based splits for the RCC dataset~\cite{Finlayson2017}. The images are gamma adjusted for visualization.}
    \label{fig:gan_indoor_rec}
\end{figure}

\subsection{Scene Based Performance}
In this experiment, the performance of \emph{CycleGAN} is evaluated based on the type of the scene, i.e. indoor vs. outdoor. The model is trained on the complete dataset, and tested on indoor and outdoor scenes separately. We partition the RECommended Color Checker (RCC)~\cite{Finlayson2017} dataset test split into indoors and outdoors. Then, we evaluate the partitions by training \emph{CycleGAN} model on the RCC training split. Table~\ref{tab:gan_indoor_rec} shows the quantitative results and Figure~\ref{fig:gan_indoor_rec} presents a number of examples. Results show that the performance of the outdoor scenes are significantly better than the indoor scenes. We believe that this is because of the more uniform nature of the lighting conditions in outdoor scenes. On the other hand, indoor scenes include more complex lighting conditions with multiple illuminants. This poses a problem as the network assumes a single light source. 

\begin{table}[h]
\centering
\scalebox{0.85}{
\begin{tabular}{|l|c|c|c|c|c|c|}
\hline
Scene & Mean & Med. & Tri. & Best 25\% & Worst 25\% & Max \\
\hline \hline
Indoor & 4.4 & 3.8 & 3.8 & 1.5 & 8.5 & 17.2\\ \hline
Outdoor & \textbf{2.4} & \textbf{1.6} & \textbf{1.9} & \textbf{0.5} & \textbf{5.3} & \textbf{8.5}\\
\hline
\end{tabular}}
\vspace{+1mm}
\caption{Performance of CycleGAN~\cite{Zhu2017} on indoor and outdoor scene of ColorChecker RECommended dataset~\cite{Finlayson2017}. It performs better on outdoor scenes where the light is more uniform.}
\label{tab:gan_indoor_rec}
\end{table}

\subsection{Comparison to the State-of-the-Art}
In this section, we compare our models with different benchmarking algorithms. We provide results for Grey-World~\cite{Buchsbaum1980}, White-Patch~\cite{Land1971}, Shades-of-Grey~\cite{Finlayson2004}, General Grey-World~\cite{Barnard2002}, First-order Grey-Edge~\cite{Weijer2007} and Second-order Grey-Edge~\cite{Weijer2007}. The parameter values $n,p,\gamma$ are set as described in~\cite{Weijer2007}. In addition, comparisons with Pixel-based Gamut~\cite{Barnard2000}, Intersection-based Gamut~\cite{Barnard2000}, Edge-based Gamut~\cite{Barnard2000}, Spatial Correlations\cite{Chakrabarti2012} and Natural Image Statistics~\cite{Gijsenij2010} are provided. Further, results for High Level Visual Information~\cite{Weijer2007-2} (bottom-up, top-down, and their combination), Bayesian Color Constancy~\cite{Gehler2008}, Automatic Color Constancy Algorithm Selection~\cite{Bianco2010}, Automatic Algorithm Combination~\cite{Bianco2010}, Exemplar-Based Color Constancy\cite{Joze2014} and Color Tiger~\cite{Banic2018} are presented. Finally, we provide comparisons for 2 convolutional approaches; Deep Color Constancy~\cite{Bianco2015} and Fast Fourier Color Constancy~\cite{Barron2016}. Some of the results were taken from related papers, thus resulting in missing entries for some datasets.

~\cref{tab:external_grayball,tab:external_cc_re,tab:external_cube} provide quantitative results for 3 different benchmarks. For the SFU Gray Ball~\cite{Ciurea2003} dataset, Table~\ref{tab:external_grayball}, Pix2Pix~\cite{Isola2017} model achieves the best performance in all the metrics with 17.5\% improvement in mean angular error, 18.4\% in median and 20.6\% in trimean. For the ColorChecker RECommended~\cite{Finlayson2017}, Table~\ref{tab:external_cc_re}, CycleGAN~\cite{Zhu2017} framework achieves the second best place in different metrics, only worse than Fast Fourier Color Constancy~\cite{Barron2016}. For the Cube dataset~\cite{Banic2018}, Table~\ref{tab:external_cube}, CycleGAN~\cite{Zhu2017} achieves the best performance in the mean error and worst and comparable results with all other methods. 

\begin{table}
\centering
\scalebox{0.68}{
\begin{tabular}{|l|c|c|c|c|c|c|}
\hline
Method & Mean  & Med. & Tri. & Best 25\% & Worst 25\% & Max \\
\hline \hline
Grey-World~\cite{Buchsbaum1980} & 13.0 & 11.0 & 11.5 & 3.1 & 26.0 & 63.0\\
Edge-based Gamut~\cite{Barnard2000} & 12.8 & 10.9 & 11.4 & 3.6 & 25.0 & 58.3\\
White-Patch~\cite{Land1971} & 12.7 & 10.5 & 11.3 & 2.5 & 26.2 & 46.5\\
Spatial Correlations~\cite{Chakrabarti2012} & 12.7 & 10.8 & 11.5 & 2.4 & 26.1 & 41.2\\
Pixel-based Gamut~\cite{Barnard2000} & 11.8 & 8.9 & 10.0 & 2.8 & 24.9 & 49.0\\
Intersection-based Gamut~\cite{Barnard2000} & 11.8 & 8.9 & 10.0 & 2.8 & 24.9 & 47.5\\
Shades-of-Grey~\cite{Finlayson2004} & 11.5 & 9.8 & 10.2 & 3.5 & 22.4 & 57.2\\
General Grey-World~\cite{Barnard2002} & 11.5 & 9.8 & 10.2 & 3.5 & 22.4 & 57.2\\
Second-order Grey-Edge~\cite{Weijer2007} & 10.7 & 9.0 & 9.4 & 3.2 & 20.9 & 56.0\\
First-order Grey-Edge~\cite{Weijer2007} & 10.6 & 8.8 & 9.2 & 3.0 & 21.1 & 58.4\\
Top-down~\cite{Weijer2007-2} & 10.2 & 8.2 & 8.7 & 2.6 & 21.2 & 63.0\\
Bottom-up~\cite{Weijer2007-2} & 10.0 & 8.0 & 8.5 & 2.3 & 21.1 & 58.5\\
Bottom-up \& Top-down~\cite{Weijer2007-2} & 9.7 & 7.7 & 8.2 & 2.3 & 20.6 & 60.0\\
Natural Image Statistics~\cite{Gijsenij2010} & 9.9 & 7.7 & 8.3 & 2.4 & 20.8 & 56.1\\
E. B. Color Constancy~\cite{Joze2014} & 8.0 & 6.5 & 6.8 & 2.0 & 16.6 &53.6\\
\hline
Pix2Pix~\cite{Isola2017} & \textbf{6.6} & \textbf{5.3} & \textbf{5.4} & \textbf{1.4} & \textbf{14.2} & \textbf{36.0}\\
CycleGAN~\cite{Zhu2017} & 8.4 & 5.9 & 6.4 & 1.5 & 19.6 & 37.8\\
StarGAN~\cite{Choi2018} & 11.4 & 9.2 & 10.0 & 3.8 & 21.8 & 41.4\\
\hline
\end{tabular}}
\vspace{+1mm}
\caption{Performance on SFU Gray Ball~\cite{Ciurea2003}.}
\label{tab:external_grayball}
\end{table}

\begin{table}
\centering
\scalebox{0.68}{
\begin{tabular}{|l|c|c|c|c|c|c|}
\hline
Method & Mean  & Med. & Tri. & Best 25\% & Worst 25\% & Max \\
\hline \hline
Grey-World~\cite{Buchsbaum1980} & 9.7 & 10.0 & 10.0 & 5.0 & 13.7 & 24.8\\
White-Patch~\cite{Land1971} & 9.1 & 6.7 & 7.8 & 2.2 & 18.9 & 43.0\\
Shades-of-Grey~\cite{Finlayson2004} & 7.3 & 6.8 & 6.9 & 2.3 & 12.8 & 22.5\\
AlexNet + SVR~\cite{Bianco2015} & 7.0 & 5.3 & 5.7 & 2.9 & 14.0 & 29.1\\
General Grey-World~\cite{Barnard2002} & 6.6 & 5.9 & 6.1 & 2.0 & 12.4 & 23.0\\
CART-based Selection~\cite{Bianco2010} & 6.1 & 5.1 & 5.3 & 2.0 & 12.1 & 24.7\\
CART-based Combination~\cite{Bianco2010} & 6.0 & 5.5 & 5.7 & 2.6 & 10.3 & 17.7\\
Pixel-based Gamut~\cite{Barnard2000} & 6.0 & 4.4 & 4.9 & 1.7 & 12.9 & 25.3\\
Intersection-based Gamut~\cite{Barnard2000} & 6.0 & 4.4 & 4.9 & 1.7 & 12.8 & 26.3\\
Top-down~\cite{Weijer2007-2} & 6.0 & 4.6 & 5.0 & 2.3 & 10.2 & 25.2\\
Spatial Correlations~\cite{Chakrabarti2012} & 5.7 & 4.8 & 5.1 & 1.9 & 10.9 & 18.2\\
Bottom-up~\cite{Weijer2007-2} & 5.6 & 4.9 & 5.1 & 2.3 & 10.2 & 17.9\\
Bottom-up \& Top-down~\cite{Weijer2007-2} & 5.6 & 4.5 & 4.8 & 2.6 & 10.5 & 25.2\\
Natural Image Statistics~\cite{Gijsenij2010} & 5.6 & 4.7 & 4.9 & 1.6 & 11.3 & 30.6\\
Edge-based Gamut~\cite{Barnard2000} & 5.5 & 3.3 & 3.9 & 0.7 & 13.8 & 29.8\\
Bayesian~\cite{Gehler2008} & 5.4 & 3.8 & 4.3 & 1.6 & 11.8 & 25.5\\
E. B. Color Constancy~\cite{Joze2014} & 4.9 & 4.4 & 4.6 & 1.6 & 11.3 & 14.5\\
Deep Color Constancy~\cite{Bianco2015} & 4.6 & 3.9 & 4.2 & 2.3 & 7.9 & 14.8\\
Second-order Grey-Edge~\cite{Weijer2007} & 4.1 & 3.6 & 3.8 & 1.5 & 8.5 & 16.9\\
First-order Grey-Edge~\cite{Weijer2007} & 4.0 & 3.1 & 3.3 & 1.4 & 8.4 & 20.6\\
FFCC (model Q)~\cite{Barron2016} & \textbf{2.0} & \textbf{1.1} & \textbf{1.4} & \textbf{0.3} & \textbf{4.6} & 25.0\\
\hline
Pix2Pix~\cite{Isola2017} & 3.6 & 2.8 & 3.1 & 1.2 & 7.2 & \textbf{11.3}\\
CycleGAN~\cite{Zhu2017} & 3.4 & 2.6 & 2.8 & 0.7 & 7.3 & 18.0\\
StarGAN~\cite{Choi2018} & 5.7 & 4.9 & 5.2 & 1.7 & 10.5 & 19.5\\
\hline
\end{tabular}}
\vspace{+1mm}
\caption{Performance on ColorChecker RECommended~\cite{Finlayson2017}.}
\label{tab:external_cc_re}
\end{table}

\begin{table}
\centering
\scalebox{0.75}{
\begin{tabular}{|l|c|c|c|c|c|c|}
\hline
Method & Mean  & Med. & Tri. & Best 25\% & Worst 25\% \\
\hline \hline
Grey-World~\cite{Buchsbaum1980} & 3.8 & 2.9 & 3.2 & 0.7 & 8.2\\
White-Patch~\cite{Land1971} & 6.6 & 4.5 & 5.3 & 1.2 & 15.2\\
Shades-of-Grey~\cite{Finlayson2004} & 2.6 & 1.8 & 2.9 & 0.4 & 6.2\\
General Grey-World~\cite{Barnard2002} & 2.5 & 1.6 & 1.8 & 0.4 & 6.2\\
Second-order Grey-Edge~\cite{Weijer2007} & 2.5 & 1.6 & 1.8 & 0.5 & 6.0\\
First-order Grey-Edge~\cite{Weijer2007} & 2.5 & 1.6 & 1.8 & 0.5 & 5.9\\
Color Tiger~\cite{Banic2018} & 3.0 & 2.6 & 2.7 & 0.6 & 5.9\\
Restricted Color Tiger~\cite{Banic2018} & 1.6 & \textbf{0.8} & \textbf{1.0} & \textbf{0.2} & 4.4\\
\hline
Pix2Pix~\cite{Isola2017} & 1.9 & 1.4 & 1.5 & 0.7 & 4.0\\
CycleGAN~\cite{Zhu2017} & \textbf{1.5} & 1.2 & 1.3 & 0.5 & \textbf{3.0}\\
StarGAN~\cite{Choi2018} & 4.8 & 3.9 & 4.2 & 1.9 & 8.9\\
\hline
\end{tabular}}
\vspace{+1mm}
\caption{Performance on Cube dataset\cite{Banic2018}.}
\label{tab:external_cube}
\end{table}

\section{Conclusion}
In this paper, the color constancy task has been modelled by an image-to-image translation problem where the input is an \emph{RGB} image of a scene taken under an unknown light source and the output is the color corrected (white balanced) one.

We have provided extensive experiments on 3 different state-of-the-art GAN architectures to demonstrate their (i) effectiveness on the task and (ii) generalization ability across different datasets. Finally, a thorough analysis is given of possible error contributing factors for future color constancy research.\\

\noindent \textbf{Acknowledgements:} This project was funded by the EU Horizon 2020 program No. 688007 (TrimBot2020). We would like to thank William Thong for the fruitful discussions.

{\small
\bibliographystyle{ieee}
\bibliography{egbib}
}

\end{document}